# A framework for GNSS-based solutions performance analysis in an ERTMS context

J. Marais, Univ Gustave Eiffel, COSYS-LEOST, F-59650 Villeneuve d'Ascq, France
Q. Mayolle, M. Fasquelle, IRT Railenium, F-59308 Famars, France
V. Tardif, E. Chéneau-Grehalle, SNCF, F-72100, Le Mans, France

## 1. Context - Progresses in GNSS-based solution introduction in rail applications

GNSS (Global Navigation Satellite System) is now used in most of our travels and each of our smartphone apps. Most of the usages are not safety-critical. But Europe identified GNSS for more applications and to be integrated in rail in general as part of the toolset to help railway to contribute to reduce transport carbon footprint. To increase the use of trains in European transports, railways must improve their attractiveness for passengers and freight, but also increase reliability, availability and efficiency by reducing capital expenditure and operational costs. GNSS is part of the global digitalization scheme of freight that aims to offer added value to the clients knowledge of accurate time of arrival, continuous monitoring of transport conditions (temperature, humidity...).

But a major challenge will be to reach stringent applications and in particular, GNSS is today seen as a realistic and serious game changer for the future of the ERTMS (European Rail Traffic Management System).

The localisation function is today performed with both odometry and balises. Odometer provides a continuous train position in time from a reference point. But as the distance delivered by the odometer shows a growing bias with distance, due to wear and wheel sliding, the use of on-track balises allows to reduce this error. Future systems will be based on on-board localisation solutions with GNSS receivers. It will allow the development of new concepts for moving blocks, virtual coupling and automation. Its use for train integrity is also investigated.

But the environmental conditions of track and surroundings configuration, i.e, tunnels, dense urban areas or vegetation often degrade positioning performance and thus its efficiency and safety. Indeed, GNSS satellites are moving and their visibility (availability and relative position from the receiver) vary with time. Moreover, for optimal performance, the system requires open sky environments, which are the cases of most of the aeronautical uses but not of train uses. Trains often circulate in areas where signal reception can be disturbed (multipath, intentional or unintentional interferences) and thus, performances degraded. If many progresses have been made in the past years to develop more robust receivers [Puccitelli, 2022], multi-sensor solutions [CLUG website] or missing tools such as Digital Maps [Crespillo, 2023], in projects such as the Shift2Rail Project X2Rail-5 or CLUG, some questions remain and in particular related to performance evaluation. How can we evaluate performances in a dynamic environment (train, satellite, obstacles)? How can we be sure that every configuration has been tested? What is the impact of a failure (inaccuracy, missed detection) on operation?

Some of these issues are addressed in the on-going R2DATO project funded by Europe's rail.

## 2. The R2DATO project and our contribution

R2DATO is a project of the Flagship Project 2 (FP2) funded under Europe's Rail. The aim of R2DATO is to take the major opportunity offered by digitization and automation of rail operation and to develop the Next Generation ATC and deliver scalable automation in train operations, up to GoA4 for 2030, to enhance infrastructure capacity on the existing rail networks. The project is coordinated by SNCF.

The FP2 R2DATO project is developing technologies in several fields of digital automated up to autonomous train operations, seeking a new paradigm in how the rail system is operated, increasing safety, flexibility, capacity, performance and reducing energy consumption and costs. The development of positioning technologies and their evaluation are part of the scope. Railenium is contributing to several tasks for SNCF. This paper will develop the concept investigated in WP34 and 35 devoted to Testing, Validation and Certification and respectively on Test Specification and architecture, then Implementation and Certification.

2.1. Contributions to the development of a HIL testbed for the evaluation of GNSS-based solution impact in an ERTMS testbed

As the performance of satellite-based positioning solutions varies over time and space, it is not possible to exhaustively demonstrate the performance of an on-board solution through long and costly test campaigns. Instead, this variety of scenarios can be carried out on a test bench, equipped with tools for simulating realistic signal reception conditions and sensor errors. However, today, no realistic models of these errors exist for railway environments. The challenge here is then to model them considering variations of the track surroundings and satellite positions in time.

GNSS errors are classically divided into global and local errors. Global errors are created by propagation through the atmosphere and by the system itself (orbit, clock, etc.). They are generally known and modelled, which means they can be at least partially corrected. Commercial signal simulators incorporate them. Local errors are, by definition, closely linked to the propagation environment close to the receiver and its antenna. They are therefore difficult to model. To illustrate this difficulty, let's take the example of the urban environment, widely covered in the literature due to the number of users and applications that are useful there. Signal propagation in a very dense urban environment, such as US or Japanese business districts, will differ from conventional European city centres, due to differences in building heights, street widths, presence or absence of vegetation, etc. In the Gate4Rail project, initial error models have been specifically defined for the railway environment, based on measurement campaigns [Gate4rail, 2020]. A few representative environments have been proposed (open sky, urban, forest), as well as the crossing of a few special features such as bridges and tunnels. However, these models are limited. They represent only partially the conditions encountered and are variable over time [Kazim, 2021].

With the development of Machine Learning techniques, the possibility of characterizing and classifying the receiving environment from GNSS observations appears to be interesting to characterize a line. These tools have been tested on the classification of indoor/outdoor or urban/open-sky environments [Gao, 2018], sometimes adding classes for trees and urban canyons [Feriol, 2022]. Limitations of these studies are: the learning databases are limited and unrepresentative of the railway environment; the number and choice of environment classes are also unsuitable. Finally, the error modelling step for each of the classes does not exist. Using railway data, [Tan, 2019] chose to classify environments along a Qinghai-Tibet line into three classes based on a visibility criterion: open sky/partial occlusion/severe occlusion. [Sun, 2022] uses classes closer to those of Gate4rail: open sky/urban/bridge/tunnel, based on a fairly short database acquired at the experimental site of the Beijing Academy of Railway Sciences. None of these models currently includes suburban environments, or the edges of wooded lines, for example.

2.2. Simulation chain

SNCF-CIM maintains an ERTMS testbed in which hard or software solutions can be tested. Our aim is to allow GNSS-based solutions to be inserted in this chain as illustrated on fig. 1. The Stella NGC Suite developed by M3Systems is the first element of the chain, capable of generating multi-constellation and multi-frequency GNSS signals along any scenario. A hardware GNSS-based solution shall then be interfaced with the simulator and feed the ERTMS chain with a localisation information. The goal of the work presented in this paper is to allow the Stella Suite to add specific local errors as representative GNSS pseudo-range errors before use by the GNSS receiver.

In order to assess the methodology, in a further step, one will compare real data with simulation, injecting local pseudo-range errors according to representation of the environment as represented in fig. 2.

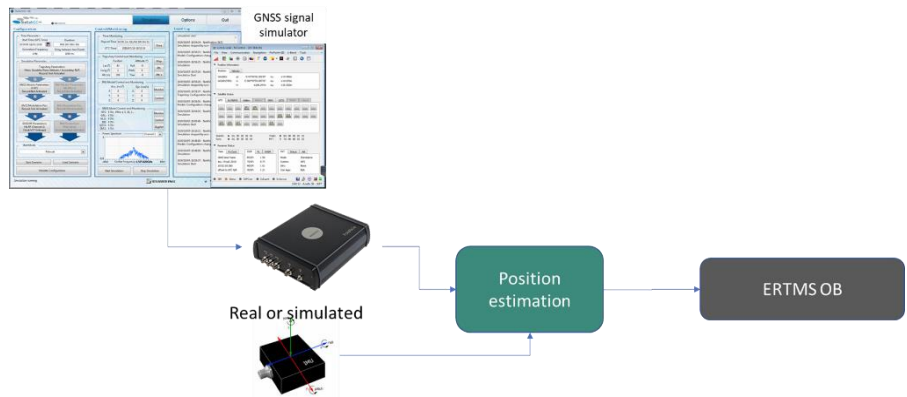

Figure 1 Schematic representation of the testing process

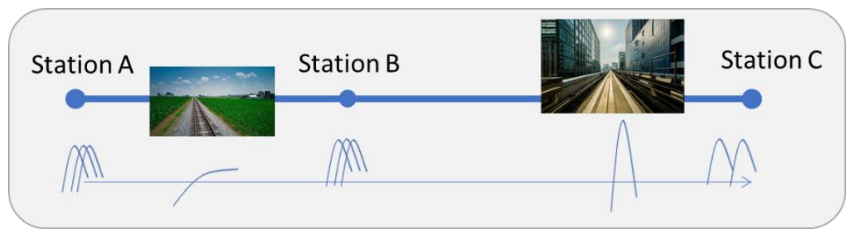

Figure 2 Local error simulation along a railway line according to the environment type detection

The first investigations done for the R2DATO project consist in:

- Proposing a data-driven context detection model
- Associating error models to each of the types of environments.

2.3. Data-driven context detection

The objective of environment classification can be twofold:

- Identify environment with semantic similarities: urban areas, streets, track with trees...
- Or identify environments with GNSS similarities: satellite visibility, pseudo-range errors...

First category is more intuitive as an operator can easily replay or describe environment and determine a series of environments encountered along a line from station A to station B. Second one may be more relevant from a GNSS view as literature still shows that "urban" for example cannot be described by a single error model anywhere on the globe.
Three main categories of features can be found in the literature for environmental classification context: signal quality variables, based on the C/N0; Constellation characterization such as satellite positions or number of satellites; pseudoranges (satellite-to-receiver distance measurements).
Our models have been trained based on several datasets. These datasets must fulfil several requirements:

- to represent train movements with multiple speeds;
- to be taken from different environment, with maximum variability: urban, open-sky, with or without trees...;
- the environments must have enough continuity (the train should spend enough time in each environment);

- to have the possibility to identify the environment a posteriori.

The last presented requirement brings the major difficulty. For each time of the train journey, we need to know the environment in which the receiver evolved. A simple solution would be to have a human operator labelling the journey from the train during the acquisition. For previously recorded data, it is necessary to use external information. For the CLUG dataset used in our study, several years have passed since the recording campaign. Data are composed of RINEX observation and navigation files, as well as a ground truth file.

A large set of Machine algorithms have been tested and compared. Only an example will be given in the following for illustration purpose. Complete analysis will be published in the coming year.

Primary environment classes are defined when the track is surrounded by a homogeneous environment (both sides of the tracks share the same environmental characteristics): Trees, Buildings, Open-sky (urban), Open-sky (rural), Bridge, Post-bridge, Station, Triage, Tunnel, Post-tunnel. Besides these classes, most of the time the two sides of the tracks do not share the same properties: for instance, one side is full of trees, and the other one is totally empty. In this situation, we define secondary classes, which are mixing of the previous ones: Mixed trees and open-sky, Mixed trees and buildings, Mixed buildings and open-sky.

## 2.4. Machine Learning process for environment separation

### 2.4.1. Machine learning general principle

The Machine Learning paradigm allows the development of complex models to perform the classification task. Starting from a large set of data with labels, the algorithm is trained to recognize this data to perform afterward the inference of the label for new unknown observations. For the GNSS application introduced in this paper, we measure the ability of the trained algorithm to distinguish between the multiple environments based on the GNSS sensor measurements only. The higher the scores reached, the better the separation of the data in their relying abstract space.

### 2.4.2. Illustration on a train journey between Fenouillet and Foix

Our exercise data consist in a two hours GNSS recording (sampling frequency of 1 Hz), giving 7000 timestamps observations, based on a train commercial train journey between Fenouillet and Foix (South of France), as illustrated in fig. 3. From these points, a manual labelling has been performed, using external sources of information for classification of the environment. For the identification of areas with buildings, Google Earth has been employed. For the vegetation, the infrared satellite images are used (publicly available on the French national Geoportail website [Geoportail]). Many timestamps were attributed to mixed environment (for instance with two different environments: building from one trackside and open-sky from the other one). Among the 7000 observations, only 3000 could be attributed to clear environments (train station, forest, buildings or opens-sky). Due to stopping intervals of the train, the observations belonging to the class train station are nearly 2000. The features computed from the observation and given to the algorithm were the statistics (mean, minimum, maximum, variances, skewness and kurtosis) of the signal characteristics for each constellation and each frequency band, alongside geometric information and number of visible satellites. The training dataset (to train the algorithms) consists in 2000 observations. The remaining 1000 observations are kept, forming the test dataset, used for the assessment of the trained algorithms performances. For multiclass problem, a simple inspection criterion is the confusion matrix, which compare the true and the predicted labels (a perfect classification algorithm would have all the counting on its diagonal).

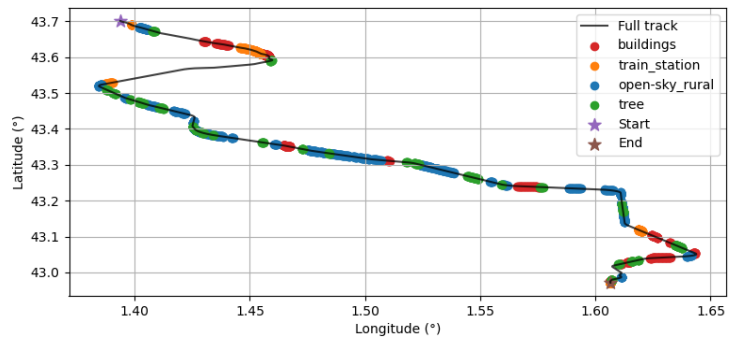

**Figure 3: Train path from the CLUG project**

2.4.3. Machine learning models: strengths and limitations

To investigate the complexity of the problem, two majors machine learning models have been employed: a linear one and a complex non-linear one. These models will provide information about the data distribution and the possible robustness of the results. In the context of machine learning, increasing the complexity of a model can bring dangerous issues, such as overfitting, which results in a model with poor generalization performances.
As a first experimentation, a linear algorithm is employed: a multi-class logistic regression (MLR) [Bishop, 2009] model. This algorithm is employed for its interpretability power, and the statistical interpretations induced by its learned parameters. To prevent overfitting, a L2 penalization is employed, whose hyperparameters are tuned by a 5-fold cross validation on the training dataset. In the case of GNSS data coming from the CLUG dataset, the MLR succeeded in identifying several environments with ease but failed for several ones, see the confusion matrix in fig. 4 (left part). This phenomenon highlights the limitations of the linear assumption. Separation between several environments is not perfect, here between the open-sky and the forest (tree) environments. This could indicate that in railways application, the vegetation around the track does not disturb necessarily the propagation of the signal. This inspection should be theoretically performed at different periods of the year to sustain this hypothesis. To fully explore the possible performances accessible by a powerful machine learning technique, the XGBoost algorithm [Chen, 2016] has been tested. This model, from the family of Gradient Boosting algorithms, is able to learn a complex non-linear separation rule, at the cost of less interpretable parameters and structure. Its empirical accuracy allows a fast estimation of the maximum performances bound, see the confusion matrix in fig. 4 (right part). The accuracy of this model is clearly superior to the previous one. This experiment shows the difficulty of the representation of environments. Various sources of information should be included to improve the separations: images from cameras, internet exchanges, or specific maps of the ground.

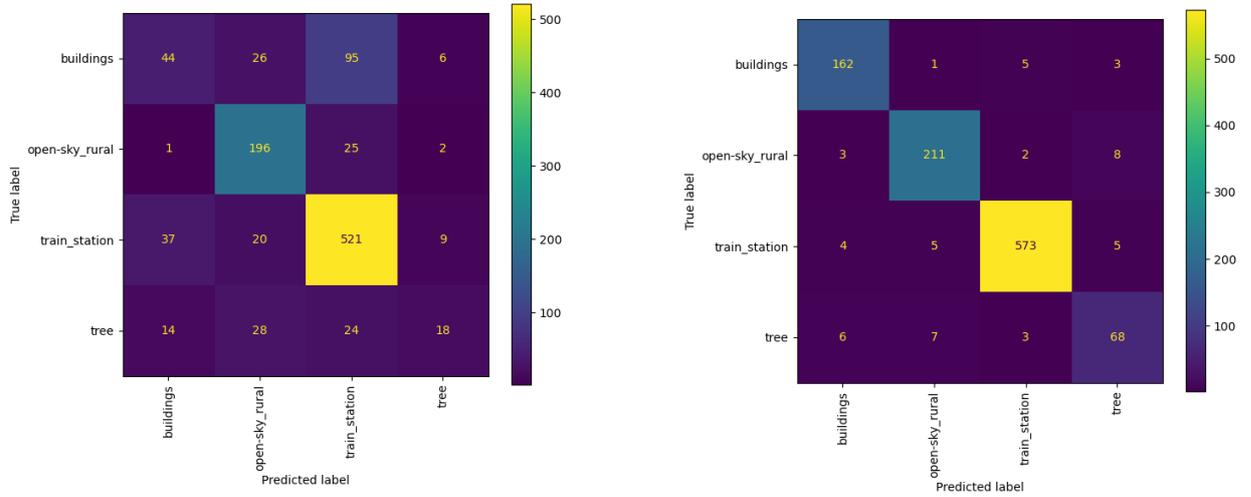

Figure 4: Confusion matrix for Multi-class Logistic Regression (left) and XGBoost (right)

The two aforementioned models are inspected after the training phase. Statistical indicators, as the SHAP value, allows the identification of the most important variables for environment separation. Among them, we discover the calculation of the ionospheric bias (therefore non-linear transformation of the elevation) and the direct minimums of RINEX observation (the statistical transformations seem not to have a high weight in the classification of environments). The other features should however not directly be removed, since the interaction between features is complex to model and their influence on the classification difficult to assess in the non-linear modelling. A general idea of the classification process is to learn the conditions of the environment. We do not want the algorithm to learn the spatial information (locations of the environments). The generalization property is indeed needed to perform classification in other areas.

2.5. Error models

The generation of a realistic local error (multipath and noise error) requires a proper knowledge of the true location of the train at each time step. This information is accessible in the CLUG dataset. From each pseudorange $R_i$, the following are sequentially removed: the geometric range $\rho\_i$, the receiver $\delta t_{rcv}$ and satellite clock offset $\delta t^{sat}$ (and relativistic correction), the instrumental, Ionospheric $I_i$ and Tropospheric delays $Tr$. Each frequency band is process independently, and the Klobuchar ionospheric model used as a standard approximation of the Ionospheric delay. The remaining quantity $\varepsilon_i$ is considered as the error to be later reproduced.

$$R_i = \rho_i + c(\delta t_{rcv} - \delta t^{sat}) + Tr + I_i + \varepsilon_i$$

The simulation we introduce is based on a stochastic model, which assumes temporal independence between errors. A Gaussian assumption of this error allows an easy parametrization of this stochastic model. The simulation therefore involves only the generation of independent samples from the Normal distribution. However, since we expect higher errors for low elevations satellites, instead of removing the related observation, this article uses a robust estimator for the parameters of the Gaussian distributions, the Minimum Covariance Determinant [Rousseeuw, 1999]. As example on the local errors calculated for the GPS satellites in the L1 frequency band is displayed in fig. 5. The robust estimator has a narrower variance than the classical estimators (empirical means and covariances): 7.7 $m^2$ instead of 19 $m^2$, since it is less biased by outlier or abnormal values.

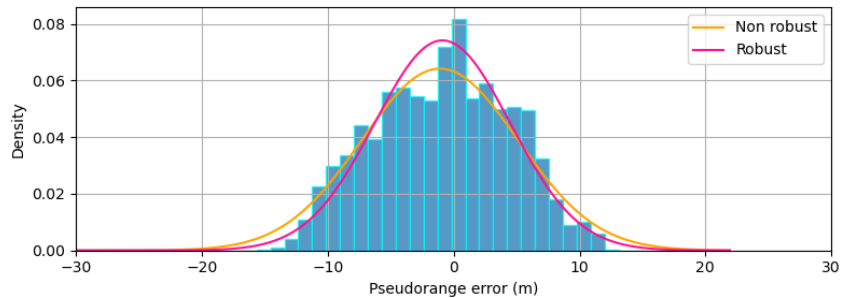

Figure 5: Local errors distributions for GPS satellites (L1) and fitted Gaussian laws

In the future, these models will be calculated for each environment, based on the empirical train journeys. Other quantities, such as those provided by the Code-Minus-Carrier method, could also be used instead of the single pseudoranges to provide more reliable estimates.

## 3. Perspectives

The work presented in this paper intends to continue the first concepts for GNSS-based solution evaluation in an ERTMS testbed initiated in the Gate4Rail project. The work is performed in a collaborative way with the CIM-SNCF, M3Systems and Railenium in order to add specific GNSS local errors in rail to the ERTMS testbed maintained by the CIM.
The concept is now in a proof-of-concept phase where the goal is to evaluate how realistic such a data-driven error model can be and if it can be used for performance demonstration and later safety evaluation. The different tasks in progress are: choice of the ML algorithm and its parameters in order to provide the best model; error generation along a railway run for simulation and evaluation; comparison analysis.

## List of Abbreviations

ERTMS: European Rail Traffic Management System
FP2: Flagship Project 2
GNSS: Global Navigation Satellite System
MLR: Multi-class Logistic Regression

## Zoom Authors

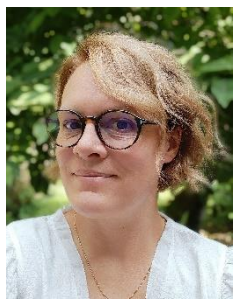
Juliette Marais received the engineering degree from ISEN and a Ph.D. degree in electronics from University of Lille, France, in 1998 and 2002 respectively. Since 2002, she has been a Researcher with INRETS, IFSTTAR and now University Gustave Eiffel. She is currently involved in GNSS performance analyses and enhancement in land transport environments and in particular in rail applications. Since 2000, she has been participating with the European Railway-Related Projects, such as the recent H2020 RAILGAP project or supporting EUSPA with expertise. Her research focuses the development of fail-safe positioning solutions for land transport applications and GNSS propagation characterization in railway environments (NLOS, Multipath, Interferences). With Railenium, she contributes to the R2DATO project.
Address: Université Gustave Eiffel, 20, rue E. Reclus, 59650 Villeneuve d'Ascq
Phone: +33.3.20.43.84.95
Email: Juliette.marais@univ-eiffel.fr

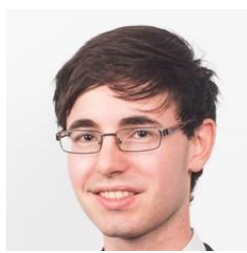
Quentin Mayolle is a Data Scientist Engineer from the École Centrale de Lille specialized in data analysis and machine learning. He received a PhD degree in 2021 and works with Railenium. He is involved in the R2DATO project.
Address: IRT Railenium, 180 Rue Joseph-Louis Lagrange, 59540 Valenciennes
Email: Quentin.mayolle@railenium.eu

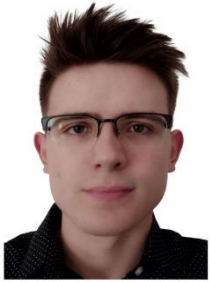
Martin Fasquelle is a Data Scientist in Railenium, from the University of Lille. He received his master degree in Applied Mathematics. He is involved in the R2DATO project.
Address: IRT Railenium, 180 Rue Joseph-Louis Lagrange, 59540 Valenciennes
Email: martin.fasquelle@railenium.eu

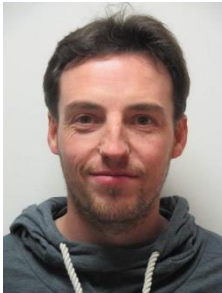
Vincent TARDIF is the tech leader from the ERTMS French Laboratory (#LEF) at SNCF Voyageurs. He received a Master degree in Sciences and Technologies from University of Le Mans in 2007. He worked as a study engineer for ALTEN until 2010 for various clients such as Orange, ERG or CEA Saclay always on software development projects. He joined the CEA Saclay then, until 2019 to work on the software development for the CIVA NDE Simulation platform. At the end of 2019, he joined the #LEF at SNCF Voyageurs, to build the software development team in order to become independent regarding the development of ERTMS test benches. In this regard, he leads the fourth task of the WP34 in the R2DATO project, which aims to define the test bench architecture for ASTP certification platforms and to build the demonstrator for this architecture.
Address: Centre d'Ingénierie du Matériel 4 Allée des Gémeaux, 72100 LE MANS
Email: vincent.tardif@sncf.fr

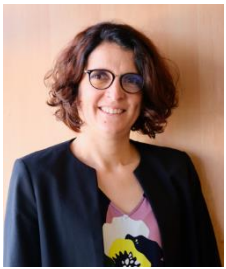
Emilie Chéneau-Gréhalle graduated from the ENSSAT engineering school, specialized in IT and electronics in 2002. After several years in telecom industries, she joined SNCF in 2016 to manage a team in charge of onboard safety equipments integration and validation. Since 2022, she is leading European projects and especially R2DATO WP34/35 dedicated to "testing, validation and certification".

Address: Centre d'Ingénierie du Matériel 4 Allée des Gémeaux, 72100 LE MANS
Phone: +33 6 01 37 34 11
Email: emilie.cheneau@sncf.frPhoto

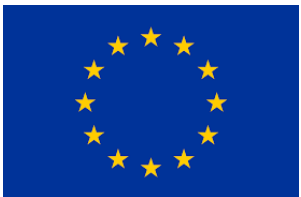
Funded by the European Union

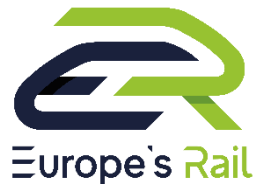
The project is supported by the Europe's Rail Joint Undertaking and its members. Funded by the European Union. Views and opinions expressed are however those of the author(s) only and do not necessarily reflect those of the European Union or the Europe's Rail Joint Undertaking. Neither the European Union nor the granting authority can be held responsible for them.